\documentclass[sigconf]{acmart}
\AtBeginDocument{%
  \providecommand\BibTeX{{%
    \normalfont B\kern-0.5em{\scshape i\kern-0.25em b}\kern-0.8em\TeX}}}

\copyrightyear{2022}
\acmYear{2022}
\setcopyright{acmcopyright}\acmConference[WWW '22]{Proceedings of the ACM Web Conference 2022}{April 25--29, 2022}{Virtual Event, Lyon, France}
\acmBooktitle{Proceedings of the ACM Web Conference 2022 (WWW '22), April 25--29, 2022, Virtual Event, Lyon, France}
\acmPrice{15.00}
\acmDOI{10.1145/3485447.3512045}
\acmISBN{978-1-4503-9096-5/22/04}

\acmSubmissionID{2695}

\newcommand{\cell}[1]{\begin{tabular}{@{}l@{}}#1\end{tabular}}

\usepackage{algorithm,algpseudocode}

\def\NoNumber#1{{\def\alglinenumber##1{}\State #1}\addtocounter{ALG@line}{-1}}

\let\oldnl\nl
\newcommand{\nonl}{\renewcommand{\nl}{\let\nl\oldnl}}

\usepackage[utf8]{inputenc} 

\usepackage{hyperref}       
\usepackage{multirow}       

\usepackage{bm}      

\usepackage{graphicx}

\newcommand{\edits}[1]{\textcolor{black}{#1}}
\DeclareMathAlphabet{\pazocal}{OMS}{zplm}{m}{n} 
\usepackage{mathtools}
\DeclarePairedDelimiterX{\norm}[1]{\lVert}{\rVert}{#1}
\usepackage{multicol}

\newcommand*{\thead}[1]{%
\multicolumn{1}{c}{\bfseries\begin{tabular}{@{}c@{}}#1\end{tabular}}}

\usepackage{amsfonts} 
\usepackage{amsmath} 
\DeclareMathAlphabet{\pazocal}{OMS}{zplm}{m}{n} 

\begin{document}
\title{Interpreting BERT-based Text Similarity via Activation and Saliency Maps}


\author{Itzik Malkiel}
\authornote{Both authors contributed equally to this research.}
\affiliation{%
  \institution{Microsoft}
  \institution{Tel-Aviv University}
  \country{Israel}
}

\author{Dvir Ginzburg}
\authornotemark[1]
\affiliation{%
  \institution{Microsoft}
  \institution{Tel-Aviv University}
  \country{Israel}}

\author{Oren Barkan}
\affiliation{%
  \institution{The Open University}
  \institution{Microsoft}
  \country{Israel}}

\author{Avi Caciularu}
\affiliation{%
 \institution{Bar-Ilan University}
 \country{Israel}}
 
\author{Jonathan Weill}
\affiliation{%
  \institution{Microsoft}
  \country{Israel}}

\author{Noam Koenigstein}
\affiliation{%
  \institution{Microsoft}
  \institution{Tel-Aviv University}
    \country{Israel}}



\begin{abstract}

Recently, there has been growing interest in the ability of Transformer-based models to produce meaningful embeddings of text with several applications, such as text similarity. Despite significant progress in the field, the explanations for similarity predictions remain challenging, especially in unsupervised settings. In this work, we present an unsupervised technique for explaining paragraph similarities inferred by pre-trained BERT models. By looking at a pair of paragraphs, our technique identifies important words that dictate each paragraph's semantics, matches between the words in both paragraphs, and retrieves the most important pairs that explain the similarity between the two. 
The method, which has been assessed by extensive human evaluations and demonstrated on datasets comprising long and complex paragraphs, has shown great promise, providing accurate interpretations that correlate better with human perceptions.

\end{abstract}

\maketitle

\section{Introduction}

The increasing availability of artificial intelligence systems has brought about an increasing demand for methods to intuitively explain machine inference and predictions \cite{selbst2018intuitive,doshi2017towards,arya2019one}. Machine explanations are utilized for a multiplicity of applications. From increasing users' trust \cite{ribeiro2016should,gilpin2018explaining, barkan2020explainable} and improving interpretability \cite{doshi2017towards, chefer2021transformer,chefer2021generic,gam,gradsam} to debugging \cite{kulesza2010explanatory}. 
In some cases, explanations are used to enhance the autonomy of people and facilitate direct oversight of algorithms \cite{barocas2020hidden}. As a result, some countries have regulations that make explanations mandatory in sensitive cases \cite{goodman2016eu,selbst2018meaningful}.

Transformer-based models have revolutionized the fields of natural language processing~\cite{vaswani2017attention,devlin-etal-2019-bert,liu2019roberta} and recommender systems~\cite{malkiel-etal-2020-recobert,bert4rec}, improving upon other neural embedding models~\cite{mikolov2013distributed, barkan2016item2vec, pennington2014glove, barkan2017bayesian, salakhutdinov2008bayesian, barkan-etal-2020-bayesian,vbn,nam,CB2CF,barkan2020cold,barkan2016modelling}. 
Significant strides were made in tasks such as machine translation \cite{vaswani2017attention}, sentiment analysis \cite{xu2019bert}, 
semantic textual similarity \cite{devlin2019bert,  malkiel-wolf-2021-maximal, barkan2020scalable}, and item similarity~\cite{malkiel-etal-2020-recobert, ginzburg-etal-2021-self,barkan2020attentive,barkan2021cold,barkan2021coldicdm}. However, transformers employ a complex attention-based architecture comprising hundreds of millions of parameters that cannot be decomposed into smaller more interpretable components. 
Hence, explaining transformer-based models is still an open question.

In this paper, we present BTI - a novel technique for explaining unlabeled paragraph similarities inferred by a pre-trained BERT model.
BTI produces interpretable explanations for the similarity of two textual paragraphs. First, both paragraphs are propagated through a pre-trained BERT model, yielding contextual paragraph representations. A similarity score such as the cosine similarity is used to measure the affinity between the paragraphs. Gradient maps for the first paragraph's embeddings are calculated w.r.t. the similarity to the second paragraph. These gradient maps are scaled by multiplication with the corresponding activation maps and summed across the feature dimensions to produce a saliency score for every token in the first paragraph. The token saliency scores are then aggregated to words, yielding word scores.

Next, BTI performs the same procedure, with the first paragraph and second paragraphs reversed. This yields word scores for the second paragraph, calculated w.r.t to the similarity with the first. Finally, the algorithm matches word-pairs from both paragraphs, scoring each pair by the importance scores associated with its elements and the similarity score associated with the pair. The most important word-pairs are then detected and retrieved as explanations.

Our contributions are as follows: (1) we present BTI, a novel method for interpreting paragraph similarity in unlabeled settings. (2) we show the effectiveness of BTI in explaining text similarity on two datasets and demonstrate its ability to pass a common sanity test. (3) we compare BTI with other alternatives via extensive human evaluations, showcasing its ability to better correlate with human perception.

\section{Related Work}
A seminal breakthrough in visualization and interpretation of deep neural networks was presented in \cite{deconvnet}. In this work, the authors proposed to employ the deconvolution operations \cite{zeiler2011adaptive} on the activation maps w.r.t. the input image, generating visual features to interpret model predictions.

A different approach was employed by \cite{backprop1,backprop2} where guided backpropagation (GBP) was proposed as an explanation method that is not restricted to visual classification models. GBP visualizes the output prediction by propagating the gradients through the model and suppressing all negative gradients along the backward pass, resulting in saliency maps of the most influenced input parts w.r.t. the downstream task.

To test the reliability of common explainability methods, \citet{adebayo2018sanity} proposed the parameter randomization test. The test suggests propagating the same input through a pre-trained and randomly initialized network to estimate the method's dependence on the underlying network weights. The proposed sanity check reveals that most of the methods described above do not pass the test, i.e., they are independent of the model parameters and therefore are not adequate for providing satisfactory model explanations. In Section 4.1, we show that BTI passes those important sanity checks.

Based on these canonical works, the NLP community introduced interpretability tools for architectures other than CNNs. The first works that proposed interpretation of textual-based networks include ~\cite{lei2016rationalizing}, and ~\cite{Ribeiro2016}. In ~\cite{lei2016rationalizing}, the authors offer to justify model predictions by generating important fragments of the inputs. Their technique utilizes explanations during training, integrating an encoder-generator paradigm that can be used to predict explanations during inference. In ~\cite{Ribeiro2016} the authors propose a ``black-box'' classifier interpretability using local perturbation to the input sequence to explain the prediction. However, since both methods rely on supervision, they can not be employed in unlabeled settings.

Another vast problem of explainability methods for NLP tasks is the variation of the input length. To adhere to this problem, the authors in ~\cite{zhao2017generating} use a Variational Autoencoder for the controlled generation of input perturbations, finding the input-output token pairs with the highest variation. While several authors adopted the sequence-pair explainability scheme, they only tackle input-output relations, hence incompatible for unlabeled text-similarity tasks.

Since the emergence of the transformer architecture ~\cite{vaswani2017attention}, several groups have tried to offer interpretation techniques for the models. exBERT \cite{hoover2019exbert}, for example, published a visual analysis tool for the contextual representation of each token based on the attention weights of the transformer model. AllenNLP Interpret \cite{wallace2019allennlp} offers token-level reasoning for BERT-based classification tasks such as sentiment analysis and NER using published techniques such as HotFlip~\cite{ebrahimi2018hotflip}, Input Reduction ~\cite{inputreduction} or vanilla gradients analysis. 

In integrated gradients (IG) \cite{sundararajan2017axiomatic}, the authors propose an explainability method that approximates an integral of gradients of the model's output w.r.t. the input. To this end, IG generates a sequence of samples by interpolating the input features with 0, computing gradients for each sample in the sequence, and approximating an integral over the produced gradients. The integral is then used to score the importance of each input feature. The authors of SmoothGrad proposed a similar yet simpler technique, by feed-forwarding noisy versions of the input, and averaging the gradients of the different outputs with respect to the objective function.

\begin{algorithm*}[t] 
\caption{BTI algorithm. The MeanShift function returns the cluster index for each sample. The clusters are sorted in descending order w.r.t. to their centroids (i.e. the top-performing cluster receives the index 0). Params: $top\_k = 2$, $p_1,p_2 \in P$.
}
\label{Algo}

\setlength{\columnsep}{+1cm}
\begin{multicols}{2}
\begin{algorithmic}[1]

\NoNumber{
\Function{MatchWords}{$\widehat{w^1},\widehat{w^2}, w^1_s, w^2_s, w^1, w^2$, reversed}
\State $M := \left\{\right\}$
\For{$1 \leq i \leq |\widehat{w^1}| $} 
    \State $m, c \gets 1,  C(\widehat{w^1_i}, \widehat{w^2_1})$ 
    \For{$2 \leq j \leq |\widehat{w^2}| $}            
    \State $\widehat{c} \gets C(\widehat{w^1_i}, \widehat{w^2_j})$ 
    \State  if $c \leq \widehat{c}$: $m, c \gets j, \widehat{c}$ 
    \EndFor
\If {reversed} \State {$M \gets M \cup \left\{\left((w^2_m, w^1_i), (w^1_s)_i \cdot (w^2_s)_m \cdot c\right)\right\}$}  \Else
\State {$M \gets M \cup \left\{\left((w^1_i, w^2_j),  (w^1_s)_i \cdot (w^2_s)_m \cdot c\right)\right\}$}
\EndIf
\EndFor
\State \textbf{return} $M$
\EndFunction   
}

\setcounter{ALG@line}{0}

\State $s^1 := (s^1_j)_{j=1}^{m} \gets \mathrm{TokenSaliency}(p_2, p_1)$

\State $s^2 := (s^2_j)_{j=1}^{m} \gets  \mathrm{TokenSaliency}(p_1, p_2)$

\State $w^1, \widehat{w^1}, w^1_s =  \mathrm{WordPiece}^{-1}(p_1, B(p_1), s^1)$

\State $w^2, \widehat{w^2}, w^2_s = \mathrm{ WordPiece}^{-1}(p_2, B(p_2), s^2)$

\State $m^1 \gets \mathrm{MatchWords}(\widehat{w^1},\widehat{w^2}, w^1_s, w^2_s, w^1, w^2, False)$   

\State $m^2 \gets \mathrm{MatchWords}(\widehat{w^2},\widehat{w^1}, w^2_g, w^1_g, w^1, w^2, True)$ 

\State \textbf{return} $\mathrm{TopWords}(m^1, m^2, top\_k)$

\end{algorithmic}
\columnbreak
\begin{algorithmic}[1]

\NoNumber{
\Function{$F_p$}{$p$} 
\State $F_{p}\gets\frac{1}{|p|}\sum_{i=2}^{|p|+1} B(p)_i$
\State \textbf{return} $F_{p}$
\EndFunction
}
\NoNumber{
\Function{TokenSaliency}{$p_1, p_2$}
\State $g \gets \frac{\partial C\left(F_{p_1}, F_{p_2}\right)}{\partial E(p_2)}$
\State $s \gets \left[\sum_{k=1}^{h} \phi \left(E\left(p_2\right)_i \circ \left(g_i\right)\right)_k\right]_{i=1}^{|p2|}$
\State $s \gets$min-max($s$)
\State \textbf{return} $s$
\EndFunction
}

\NoNumber{
\Function{TopWords}{$m^1, m^2, top\_k$}
\State $M \gets m^1 \cup m^2$
\State $M^{*} \gets ((M_j)_2)_{j=1}^{|M|}$ \Comment{word-pair scores}
\State  $S := (c)_{k=1}^{|m^1|+|m^2|} \gets$ MeanShift$\left(M^{*}\right)$  \Comment{the cluster index for each sample}
\State $T := \left\{\right\}$
\For{$1 \leq i \leq |M| $} 
\State {$T \gets T \cup \{M_i\}$, if $S_i \leq  topC$}
\EndFor
\State \textbf{return} $T$
\EndFunction
}
\end{algorithmic}
\end{multicols}
\end{algorithm*}

\section{Method}

In this section, we present the mathematical setup of BTI and its applications for explainable text-similarity in self-supervised pre-trained language models. 
We build our method upon BERT, however, BTI can be applied with other Transformer-based language models.

\subsection{Problem Setup}

Let $\mathcal{V}=\left\{t_{i}\right\}_{i=1}^{V}$ be the vocabulary of 
{\color{black}words} in a given language. Let $S$ be the set of all possible sentences induced by $\mathcal{V}$. Additionally, let $P$ be the set of all possible paragraphs generated by $S$.

BERT can be defined as a function $B: P \rightarrow \mathbb{R}^{N \times h}$, where $h$ is the hidden layer size, and $N = 512$ is the maximal sequence length supported by the model. In inference, BERT receives a paragraph $p \in P$ and decomposes it into a sequence of {\color{black}$q \in \mathbb{N}$ tokens} $(p^j)_{j=1}^{q}$, by utilizing the WordPiece model \cite{wu2016google}. 
The sequence is then wrapped and padded to $N$ elements, by adding the special $CLS$, $SEP$ and $PAD$ tokens. This token sequence can be written as $I^p := (CLS, (p^j)_{j=1}^{q}, SEP, ..., PAD)$.

In BERT, all tokens are embedded by three learned functions: token, position and segment embeddings, denoted by $T$, $O$, and $G$, respectively. The token embedding transforms tokens' unique values into intermediate vectors $T(I^p) \in \mathbb{R}^{N \times h}$. The position embedding encodes the token positions to the same space, $O(I^p) \in \mathbb{R}^{N \times h}$. The segment embedding is used to associate each token with one out of two sentences $G({\{0, 1\}}^{N}) \in \mathbb{R}^{N \times h}$ (as standard BERT models are trained by sentence pairs). 

In this work, we feed BERT with single paragraphs and leave the use of paragraph-pairs sequences to future investigation. The principle behind this choice stems from two aspects: (1) in many cases, paragraph similarity labels do not exist, therefore, we can not fine-tune the language model to the task of paragraph-pairs similarity. 
This entails the use of a pre-trained language model, that is commonly trained by sentence pairs or chunks of continuous text, and does not specialize in non-consecutive paragraph pairs. Therefore, the inference of non-consecutive paragraph-pairs may introduce instabilities, due to the discrepancy between both phases. 
(2) the technical limitation of maximal sequence length in BERT architecture, for which feeding two paragraphs as a unified sequence may exceed the limit of 512 tokens.

\begin{figure*}[t]
\includegraphics[width=0.85\linewidth]{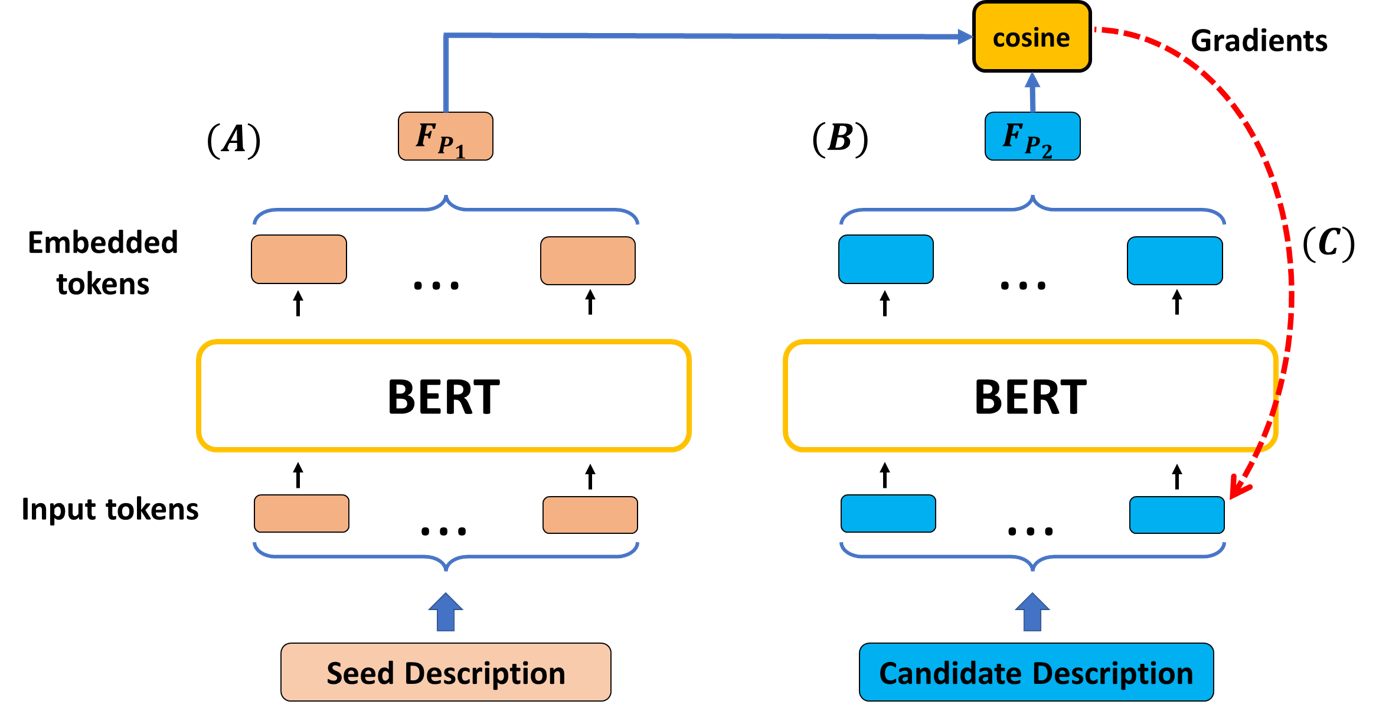}
\caption{Illustration of the gradient calculation w.r.t. the similarity between two bodies of text. (A) the description of the seed item is propagated through the model, which then outputs embeddings that are averaged pooled into a single feature vector. (B) the same procedure is applied to the description of the candidate item, where the model retrieves a feature vector that represents the candidate text. (C) a similarity function, such as the cosine similarity, is applied between both vectors, followed by a gradients calculation on the input embeddings of the candidate text w.r.t. to the similarity score. The same procedure is applied twice, where the roles of the seed and the candidate text are reversed (not shown in the figure).}  
    \label{fig:gradients_calc}
\end{figure*}

\begin{figure}[t]
\vspace{+6pt}
\includegraphics[width=1\linewidth]{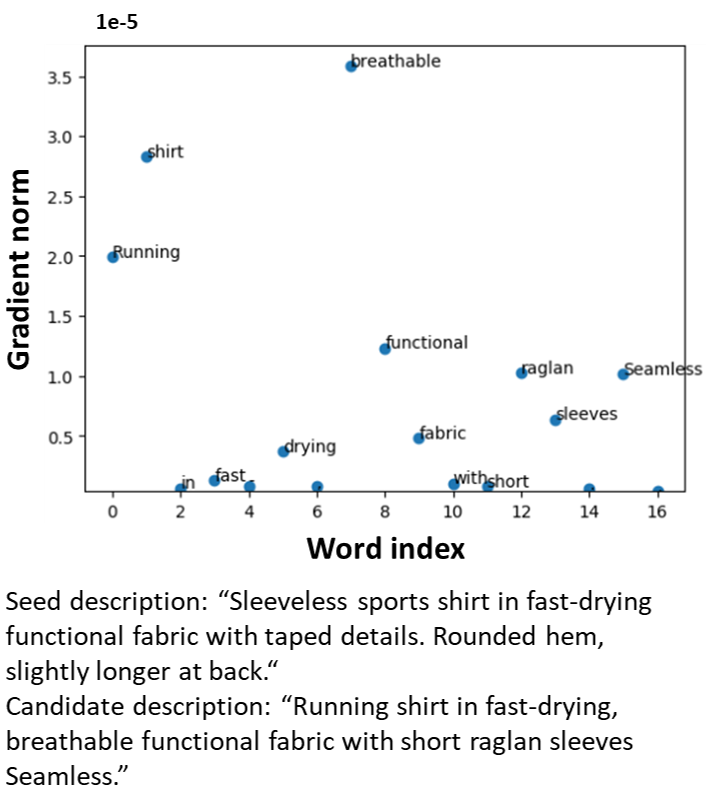}
\caption{
A visualization of the gradients norm of a representative candidate description calculated w.r.t. its cosine similarity with its seed item. The words associated with the highest gradient norm are considered to be the most important words in the candidate's textual description. The seed description and candidate description appear at the bottom of the graph. }  
    \label{fig:gradients_sample}
\end{figure}

\subsection{BERT Interpretations (BTI)}
\vspace{+4pt}

By propagating $p$ through BERT, the model produces a sequence $B(p) \in R^{N \times h}$ of latent tokens. Each element in $B(p)$ is associated with its matched element in $I^p$. A feature vector can be computed for each paragraph by average pooling the output tokens as follows:
\begin{equation}
\label{eq:feature_vector}
F_{p} := \frac{1}{q}\sum_{i=2}^{q+1} B(p)_i,
\end{equation} 
omitting the latent elements associated with the $CLS$, $SEP$ and $PAD$ input tokens.

{\color{black}
The similarity between paragraph-pairs can be interpreted by highlighting and matching important words from each element. The highlighted words in both elements should dictate the semantical similarity of the two paragraphs in a way that correlates with human perception. Hence, given two paragraphs denoted by $p_1 \in P$
and $p_2 \in P$,
we build BTI to identify important word-pairs with similar semantics} $M := {\left\{(x_i,y_i)\right\}}_{i=1}^{e}$, {\color{black}where $x_i \in p_1$ and $y_i \in p_2$, for all $i$}, and $e$ is the number of pairs detected by BTI.

Since each paragraph is fed as a separate sequence, we build our method upon the $T$ and $O$ embeddings. Specifically, given the paragraphs $p_1$ and $p_2$, BTI first calculates saliency maps, for each paragraph, by utilizing the activation map values $E(I^{p_1}) \in \mathbb{R}^{q_1 \times h}$ and $E(I^{p_2}) \in \mathbb{R}^{q_2 \times h}$ where
\begin{equation}
\label{embedding_layer}
E(I^p) := T\left({I^p}\right) + O\left((i)_{i=1}^{N}\right)  
\end{equation}
along with the gradients calculated on the same activations w.r.t. to a cosine score between the feature vectors of both paragraphs. {\color{black} $q_1$ and $q_2$ denote the number of tokens in the WordPiece decomposition of $p_1$ and $p_2$, respectively.}

The BTI algorithm is depicted in Alg.\ref{Algo}. In lines 1-2, BTI invokes the TokenSaliency function to infer a \emph{token-saliency} score for each token in the given paragraph-pair. The TokenSaliency is first applied to $(p_1, p_2)$, then, in line 2, the roles of $p_1$ and $p_2$ are reversed. Since both sides are analogue, we will describe TokenSaliency by its first application. 
{\color{black}
For a given paragraph pair $(p_1,p_2)$, the function propagates each paragraph through BERT, calculates the feature vectors $F_{p_1},F_{p_1}$ and

the partial derivations, $g(p_1,p_2) \in \mathbb{R}^{N \times q_2}$, of the input embeddings $E(I^{p_2})$ w.r.t. to the cosine between the two feature vectors $F_{p_1}$ and $F_{p_1}$ (the first feature vector is used as a constant, the second is generated by propagating $p_2$ through the model to derive gradients on its intermediate representations). Formally:
\begin{equation}
\label{eq:grads}
g(p_1,p_2) := 
\frac{\partial \mathrm{C}\left(F_{p_1}, F_{p_2}\right)}{\partial E(p_2)}
\end{equation}
where $C$ {\color{black}is a similarity function. In our experiments we utilize the cosine similarity function}\footnote{$C(v,u) = \frac{v \cdot u}{\left\Vert v \right\Vert \left\Vert u\right\Vert}$ where $v,u \in \mathbb{R}^h$}. }
{\color{black}Importantly, the positive gradients of the above term are the ones that maximize the cosine similarity between the two paragraphs (whereas stepping towards the direction of the negative gradients, would entail a smaller cosine similarity).} \edits{An illustration of this gradients calculation can be seen in Fig.~\ref{fig:gradients_calc}. For completeness, we also provide a figure demonstrating the effectiveness of the gradients, calculated on a representative sample of seed-candidate descriptions. In Fig.~\ref{fig:gradients_sample} we present the gradients' norm calculated on the words of a given candidate description w.r.t. to the similarity with its matched seed description.   }

The gradients are then multiplied by the activation maps of the same embeddings:
\begin{equation}
\label{eq:saliency}
   s = \mathrm{NRM}\left[\left(\sum_{k=1}^{h} \phi \left(E\left(p_2\right)_i \circ g_i\right)_k\right)_{i=1}^{q_2}\right]
\end{equation}
where $\phi$ is the ReLU activation function, $\circ$ is the Hadamard product, and NRM is the min-max normalization, which transforms the data to [0,1]. 

The motivation behind Eq.~\ref{eq:saliency} is as follows: we expect important tokens to have embedding values and gradients that agree in their sign -  namely both positive or both negative. 
This indicates that the gradients that stem from $C\left(F_{p_1}, F_{p_2}\right)$ emphasize the relevant parts of the embeddings that are important for the similarity between the two paragraphs. 
Additionally, embeddings and gradients with higher absolute values are more significant than those with values near zero.

\begin{figure*}[t]
\centering
\resizebox{1.0\linewidth}{!}{%
\begin{tabular}{c}

    (a) \cell{\includegraphics[width=0.98\linewidth]{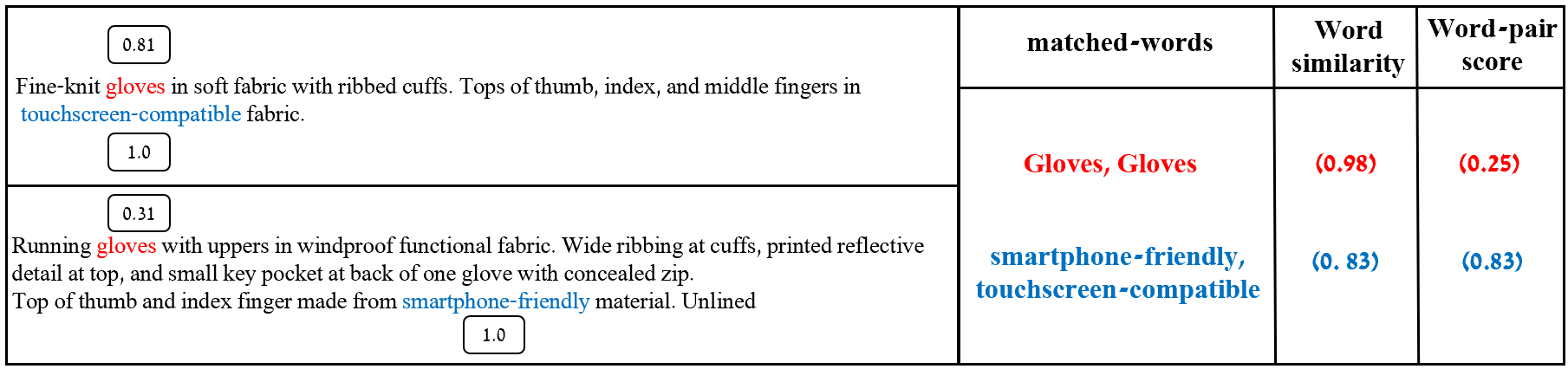}}\\
    (b) \cell{\includegraphics[width=0.98\linewidth]{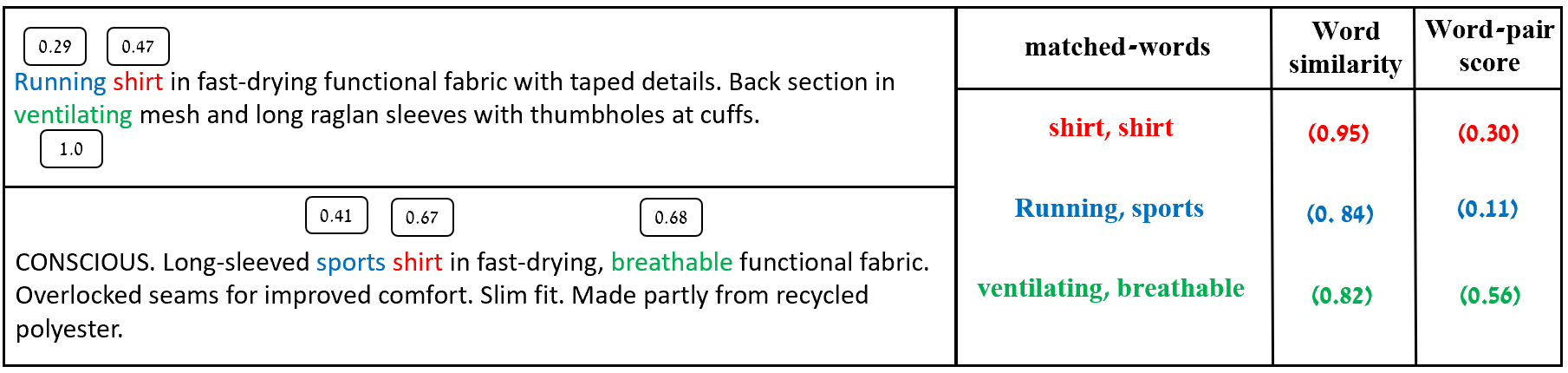}}\\
    (c) \cell{\includegraphics[width=0.98\linewidth]{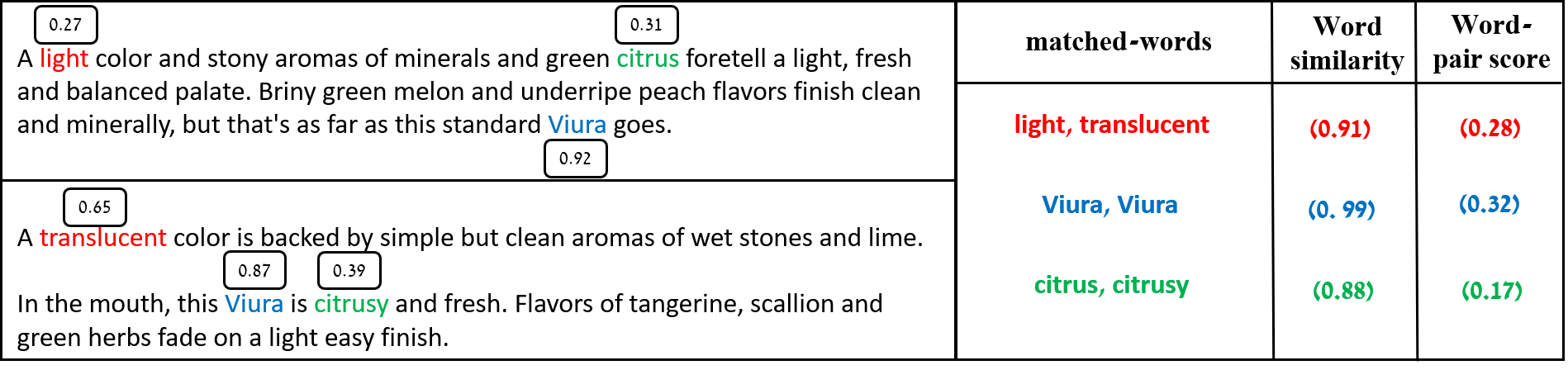}} 
\end{tabular}}
\caption{BTI results for explaining item similarities in the fashion (a-b) and wines (c) datasets. Each sample comprises two paragraphs: (1) the seed item description (top left), and (2) the description of the most similar item w.r.t. the seed (bottom left). The words retrieved by BTI are associated with saliency scores (shown next to each word). The ``matched-words'' column depicts the word-pairs chosen by BTI. The ``word similarity'' column presents the cosine scores between the word pairs. The ``word-pair score'' column exhibits the score from Eq.\ref{eq:word-pair-score}.}  
    \label{fig:fashion}
\end{figure*}

In line 3-4, BTI applies the $\mathrm{WordPiece}^{-1}$ function. This function receives (1) the tokens, (2) the latent token representations\footnote{generated by propagating the paragraph through BERT}, and (3) the token-saliency scores of a given paragraph. The function then aggregates all arguments to word-level representation, retrieving (1) whole words (rebuilt from tokens), (2) latent representation for each word, and (3) word-saliency scores. The second and third aggregations employ predefined functions $\varphi$ and $\psi$ on the latent tokens and token-saliency scores associated with the same word, respectively. The result word-level representation is then retrieved as an output. $w^1$ and $w^2$ denote the words sequences produced by aggregating the tokens of $p_1$ and $p_2$, respectively. $\widehat{w^1}$ and $\widehat{w^2}$ denote the aggregated latent word-level representation of $p_1$ and $p_2$, respectively. Analogously, the word-saliency scores are denoted by $w^1_s$ and $w^2_s$. 

In our experiments, we define $\varphi$ and $\psi$ as the mean and max functions. This entails that (a) the latent representation of a word would be defined by the mean representation of its tokens, and (b) the importance of a given word would be matched to the maximal importance of its tokens. For example, assuming a given paragraph comprising the word ``playing'', for which the BERT tokenizer decomposes the word to the tokens ``play'' and ``ing''. Assuming the TokenSaliency function assigns the tokens with the token-saliency scores 
$0.1$ and $0.8$, respectively. Then, the importance of the word ``playing'' would be associated with $0.8$.

By calling MatchWords, in lines 5-6, BTI identifies word pairs from both paragraphs that share the most similar semantics. Specifically, for each word $w^1_i \in w^1$, the function retrieves a matched word 
$w^2_j \in w^2$ that maximizes the similarity score between the aggregated latent representation of the words
\begin{equation}
w^2_{i^{*}} := \arg \max _{w^2_j \in {w^2}} \mathrm{C}\left(\widehat{w^1_i},\widehat{w^2_j}\right)
\end{equation}
where $\widehat{w^1_i} \in \widehat{w^1}$ and $\widehat{w^2_j} \in \widehat{w^2}$ are the means of the latent tokens associated with the words $w^1_i$ and $w^2_j$, respectively.

In addition to conducting matches between word pairs, the MatchWords function calculates a \emph{word-pair score} for each pair. The word-pair score represents the accumulated importance of the pair and is defined as the multiplication of the word scores of both words along with the cosine similarity between the latent representation of the words. Formally, the word-pair score of the pair ($w^1_i$, $w^2_i$) can be written as: 
\begin{equation}
\label{eq:word-pair-score}
U(w^1_i, w^2_i) := c^{ij} \cdot s^1_i \cdot s^2_j
\end{equation}
where $c^{ij}$ is the cosine similarity between the embeddings of $w^1_i$, $w^2_j$, and $s^1_i$, $s^2_j$ are the saliency scores of the words $w^1_i$ and $w^2_j$, respectively.

In line 7, BTI calls the TopWords function, which retrieves a sub-sequence of the most important word-pairs by clustering the word-pairs scores, and identifying the top-performing clusters. For retrieving the most important word-pairs, we run the MeanShift algorithm \cite{comaniciu2002mean} on the set of word-pairs scores, to obtain the modes of the underlying distribution. MeanShift is a clustering algorithm that reveals the number of clusters in a given data and retrieves the corresponding centroid for each detected cluster. In our case, MeanShift is applied to the 1D data of all $U(w^1_i, w^2_j)$ and identifies the subsets of the most important pairs, as the cluster associated with the top\_k centroids. In BTI, top\_k is a predefined hyperparameter. The detected most important word-pairs are retrieved as a sequence, which can be then visualized to interpret the similarity between the given two paragraphs.

\begin{figure*}[t]
\centering
\resizebox{1.0\linewidth}{!}{%
\begin{tabular}{c}
    (a) \cell{\includegraphics[width=0.98\linewidth]{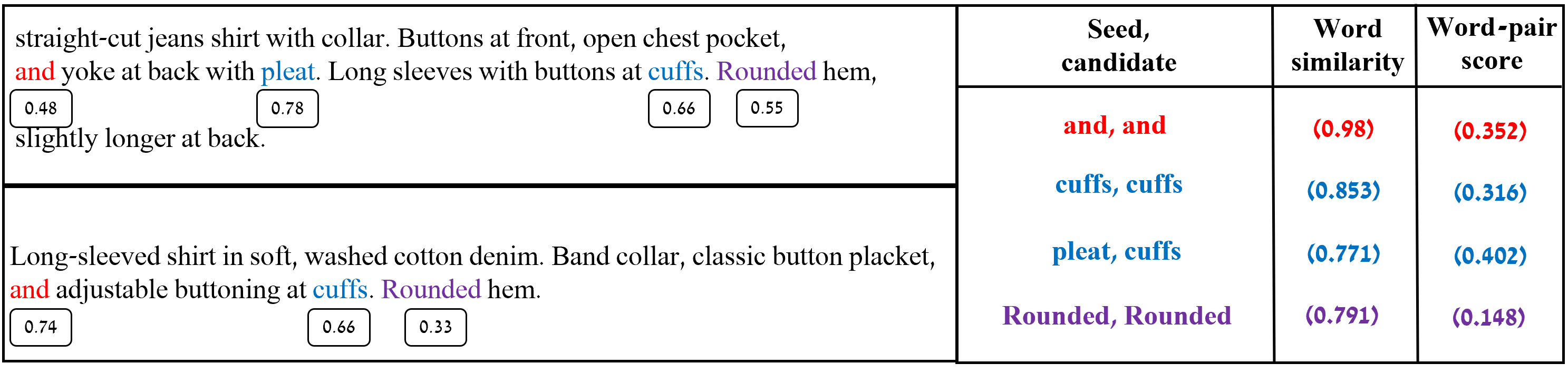}}\\
    (b) \cell{\includegraphics[width=0.98\linewidth]{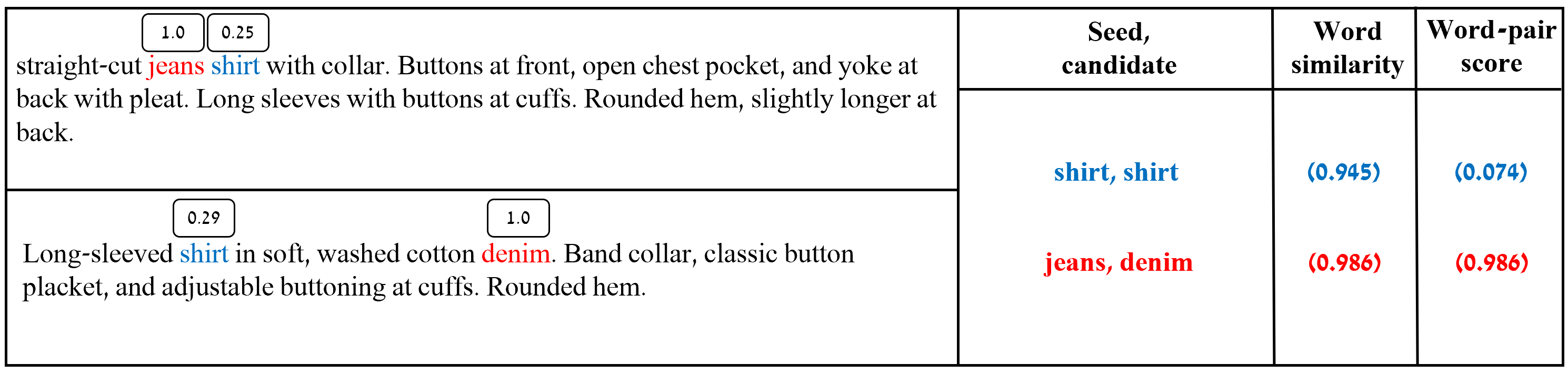}}\\
    (c) \cell{\includegraphics[width=0.98\linewidth]{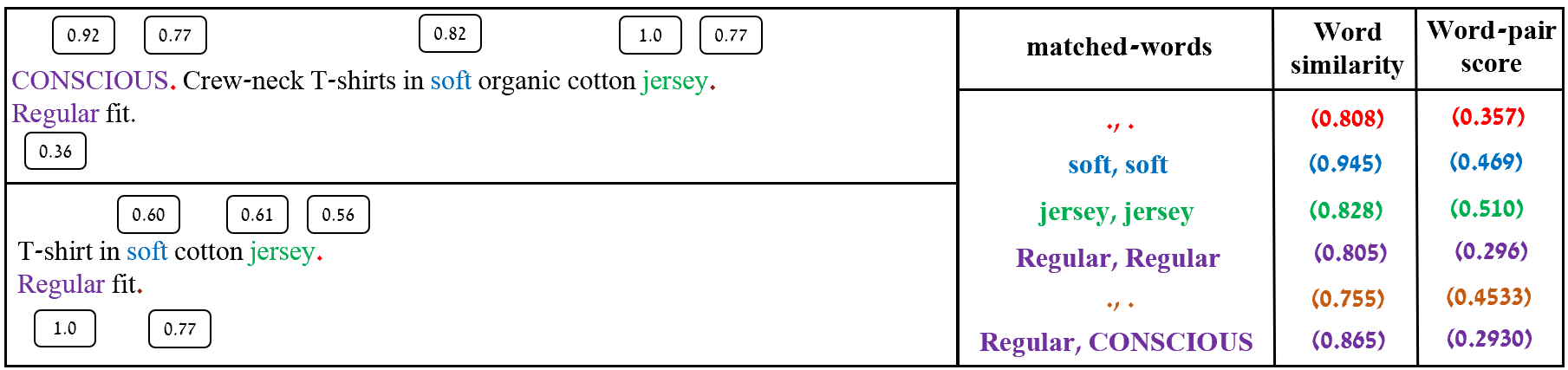}}\\
    (d) \cell{\includegraphics[width=0.98\linewidth]{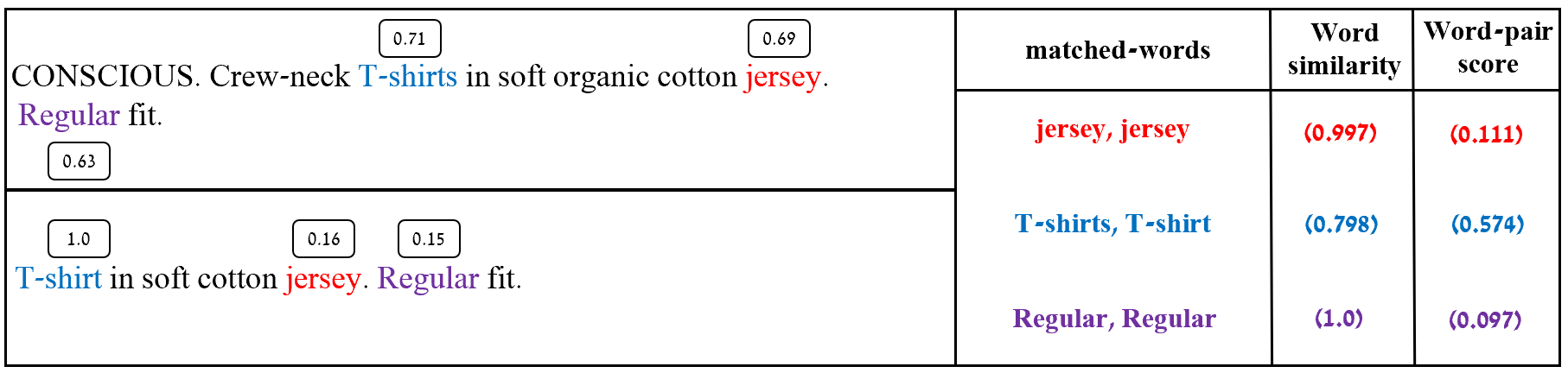}}

    \end{tabular}}
  \caption{Representative results for the parameter randomization test of BTI applied on a randomly initialized network (a,c) and BTI applied to the pre-trained weights (b,d). 
  }
    \label{fig:sanity}
\end{figure*}

\section{Experiments}
We first demonstrate BTI's ability to explain the similarity between items with similar descriptions, evaluated on a fashion and wine reviews\footnote{https://www.kaggle.com/zynicide/wine-reviews} datasets, comprising $\sim$1000 and $\sim$120K items, respectively.
The items from both datasets contain textual descriptions in the form of single paragraphs. In our experiments, we train a separate RecoBERT ~\cite{malkiel-etal-2020-recobert} model for each dataset. RecoBERT is a BERT-based model trained for item similarity by optimizing a cosine loss objective and a standard language model. \edits{The choice of RecoBERT stems from its ability to effectively score the similarity between two bodies of text, and since it does not require similarity labels for training. }

{\color{black}RecoBERT training begins by initializing the model with a pre-trained BERT model and continues training by feeding the model with title-description pairs, extracted from the items in the dataset. The training objective of RecoBERT comprises a standard masked language model (similar to BERT) along with a title-description model, which enforces the model to embed item descriptions under a well-defined metric (such as the cosine function). Specifically, RecoBERT optimizes the BERT-based backbone model to embed text pairs extracted from the same item as vectors pointing to the same direction, while text pairs sampled from different items are reinforced to have in opposite directions.} 

Given a trained model, we infer item-to-item similarities for each dataset by propagating all item descriptions through the specialized RecoBERT model. For each item, we extract a feature vector, as presented in Eq.~\ref{eq:feature_vector}. Then, given a seed item $s$, we calculate the cosine similarity between $F_{s}$ and the feature vectors of all the other items in the dataset. The \emph{candidate} item that maximizes the cosine similarity is retrieved as the \emph{most similar item}. Finally, in order to interpret the similarity between each seed and candidate item, we employ BTI on the descriptions of both. 

Figure~\ref{fig:fashion} presents the interpretations of BTI for three representative seed-candidate items from the fashion and wine datasets. In the first sample, comprising two paragraphs of two glove items, BTI highlights the words ``gloves'' in both items, which indicates the type of the item, as well as the phrases ``touchscreen-compatible'' and ``smartphone-friendly'', that strongly explain the similarity between the items. 
For the second sample, BTI highlights the category of both items (``shirt'') and other key characteristics including ``running'' and ``sports'', as well as ``ventilating'' and ``breathable'', that strongly correlate. The third sample in the figure, consists of two white wine items from the wines dataset. For this sample, BTI highlights the variety of wine grape (``Viura''), the color of both wines (``light'', ``translucent'') as well as a dominant flavor in both items (``citrus'', ``citrusy'').

By obtaining a faithful interpretation for the similarity between paragraphs that relies on intermediate representations extracted from the model, BTI reveals the reasoning process behind RecoBERT's embedding mechanism. Importantly, any use of other techniques for interpreting paragraph similarities that do not utilize RecoBERT's intermediate representations would be independent of the model weights and, therefore, will not assess the sanity of the model. This is not the case in BTI, which strongly relies on the model intermediate representations, and therefore, can assess the validity of the model. Due to the above property, BTI can allow researchers to debug their language models by analyzing the embeddings of similar paragraphs, even when similarity labels do not exist.

\label{subsec:sentence-sim}

\subsection{Sanity Test}

\begin{table}[t]
\centering
\smallskip\noindent
\resizebox{0.68\linewidth}{!}{%
\begin{tabular}{lc} 
\thead{Model} & \thead{MOS}  \\
\hline
(i) BTI (last layer)&  2.6 $\pm$ 0.5 \\
(ii) BTI (activations) &  2.4 $\pm$ 0.6\\
(iii) BTI (gradients) &  3.9 $\pm$ 0.3\\
\hline
TF-IDF-W2V & {2.5} $\pm$ 0.5 \\
IG & {3.5} $\pm$ 0.4 \\
VG & {3.6} $\pm$ 0.5 \\
BTI & \textbf{4.3} $\pm$ 0.3 \\
\hline
\end{tabular}}
\caption{Human evaluations of explainability. Shown are mean and standard deviation results over all human judges. 
}\vspace{-8mm}
\label{Tab:ablation}
\end{table}

To assess the validity of BTI for explanations, we conduct the \emph{parameter randomization} sanity test from \cite{adebayo2018sanity}. 
This test uses the explainability method twice, once with random weights and once with the pre-trained weights.

Fig.~\ref{fig:sanity} exhibits two representative samples processed twice by BTI. In the first application, BTI is applied with a randomly initialized BERT network. In the second, BTI employs the pre-trained BERT weights. In the figure, we see that when BTI utilizes the pre-trained model, it produces semantically meaningful interpretations and fails otherwise. 

Specifically, in Fig.~\ref{fig:sanity}(a), BTI utilizes BERT with random weights, identifies connectors (such as and), and fails by retrieving non-important word pairs (such as cuffs, pleats). For the same sample, as shown in Fig.~\ref{fig:sanity}(b), BTI employing the prescribed weights of the pre-trained BERT model, identifies the important words that explain the similarity between the two paragraphs by retrieving the type of fabric (denim, jeans) and the type of clothing (T-shirt), that strongly correlate both paragraphs by marking the most important mutual characteristics. 

In Fig.~\ref{fig:sanity}(c), we observe that BTI with random weights fails to identify the category of the two items (missing the fact that the two items are T-shirts), and puts attention on two dot tokens. These problems are not apparent when BTI is applied on the trained model, for which, as can be seen in Fig.~\ref{fig:sanity}(d), BTI highlights the category of the item, avoids non-important punctuations, and focus on the fabric and style of the two items.

\subsection{Human Evaluations of Explainability}

The compared baselines are Vanilla Gradients (VG), Integrated Gradients (IG) ~\cite{sundararajan2017axiomatic}, and a TF-IDF based method (denoted by TF-IDF-W2V).

\begin{itemize}
    \item Vanilla Gradients (VG) - This method interprets the gradient of the loss with respect to each token as the token importance.
    \item Integrated Gradients (IG) - This method define an input baseline $x'$, which represents an  input absent from information. We follow \cite{wallace2019allennlp} and set the sequence of all zero embeddings as the input baseline. For a given input sentence $x$, word importance is defined by the integration of gradients along the path from the baseline to the given input. IG requires multiple feed-forward to generate the explainability of each sentence.
    \item TF-IDF-W2V - As a baseline method, we consider an alternative model to the word-pair scoring from Eq.~\ref{eq:word-pair-score}:
\begin{equation*}
    \Lambda(w^1_i,w^2_j):=T(w^1_i)T(w^2_j)\text{C}(W(w^1_i),W(w^2_j))
\end{equation*}
where $T$ and $W$ are the TF-IDF scoring function and the word-to-vector mapping obtained by a pretrained word2vec (W2V) model \cite{mikolov2013distributed}, respectively. Hence, we dub this method TF-IDF-W2V. We see that $\Lambda$ incorporates both the general words \emph{importance} (captured by TF-IDF scores) as well as their semantic relatedness (captured by W2V). 
\end{itemize}

Table~\ref{Tab:ablation} presents an ablation study for BTI and a comparison with vanilla gradients (VG), integrated gradients (IG), and TF-IDF-W2V on the fashion dataset. For the ablation, the following variants of BTI are considered: 
(i) utilizing the gradients and activations of the last BERT layer instead of the first one. 
(ii) using token-saliency scores based on the activation maps alone, 
and (iii) using scores solely based on gradients. 
The last two variants eliminate the Hadamard multiplication in Eq.~\ref{eq:saliency}, utilizing one variable at a time.

To compare BTI with VG, and IG, we replace the score being differentiated by the cosine similarity score (rather than the class logit). Since the above methods are not \emph{specific}, i.e., they produce intensity scores for all words in the paragraphs, the comparison is applied via scoring the correlation of the most highlighted words, retrieved by each method, with the human perception.

To assess our full method's performance against other alternatives and compare it to the ablation variants above, we report interpretation scoring, using a 5-point-scale Mean Opinion Score (MOS), conducted by five human judges. The same test set, comprising 100 samples, was ranked for all four BTI variants (full method and three baselines) and three baselines above. The scoring was performed blindly, and the samples were randomly shuffled. Each sample interpretation was ranked on a scale of 1 to 5, indicating poor to excellent performance.

The results in Tab.~\ref{Tab:ablation}, highlight the superiority of BTI compared to other alternatives, indicating a better correlation with human perception. As for the ablation, the results indicate the importance of utilizing the gradients on the embedding layer from Eq.~\ref{embedding_layer} as described in this paper and emphasize the importance of the multiplication between gradient and activations.

\section{Summary}

We introduce the BTI method for interpreting the similarity between paragraph pairs. BTI can be applied to various natural language tasks, such as explaining text-based item recommendations.
The effectiveness of BTI is demonstrated over two datasets, comprising long and complex paragraphs. Additionally, throughout extensive human evaluations, we conclude that BTI can produce explanations that correlate better with human perception. For completeness, we show that BTI passes a sanity test, commonly used in the computer vision community, that estimates the reliability of explainability methods. 

Finally, we believe that BTI can expedite the research in the domain of language models by identifying failure modes in transformer-based language models and assessing deployable language models' reliability.

\clearpage
\bibliographystyle{ACM-Reference-Format}
\bibliography{main}
\newpage

\end{document}